\documentclass[runningheads]{llncs}
\usepackage[T1]{fontenc}
\usepackage{comment}
\usepackage{graphicx}
 \usepackage{amsmath} 
\usepackage{multirow}  
\usepackage{xcolor}
\usepackage{amssymb}
\usepackage{booktabs}
\usepackage{pifont}
\newcommand{\cmark}{\ding{51}}
\newcommand{\xmark}{\ding{55}}
\usepackage{hyperref} 
\usepackage{wrapfig}
\begin{document}
\title{FedV-KGQA: Multi-Hop Question Answering over Vertically Partitioned Knowledge Graphs}

\titlerunning{FedV-KGQA: Vertical Federated KGQA}
%
\author{Md Saikat Islam Khan Bappy\inst{1}\orcidID{0009-0009-1768-6102} \and
Oshani Seneviratne\inst{1}\orcidID{0000-0001-8518-917X}}
\authorrunning{Bappy and Seneviratne}
\institute{Rensselaer Polytechnic Institute, Troy, NY 12180, USA\\
\email{\{islamm9, senevo\}@rpi.edu}}

\maketitle              
\begin{abstract}
Real-world data for knowledge graph question answering is often distributed
across different organizations due to governance and data sovereignty constraints. While centralized systems exist, they cannot answer multi-hop questions when
the required facts are split across vertically partitioned silos. In this
paper, we propose \textbf{FedV-KGQA}, a framework for multi-hop reasoning
over knowledge graphs in which organizations share entities but own disjoint
sets of relations. Our approach combines local graph enrichment and knowledge
graph embeddings to ensure raw triples and relation parameters never leave
each silo, establishing a structural data boundary without requiring
centralized graph access. We further introduce a topic entity anchoring
mechanism that grounds questions in the correct graph neighborhood without
any runtime inter-silo communication. 
We evaluate 12 model configurations across three benchmarks and show that FedV-KGQA performs strongly, remains close to centralized performance, generalizes to 3-hop reasoning, and is robust to embedding perturbations.


\end{abstract}
\section{Introduction}
\label{sec:intro}

\begin{figure}[t]
    \centering
    \includegraphics[
        width=\linewidth,
        height=0.33\textheight,
        keepaspectratio
    ]{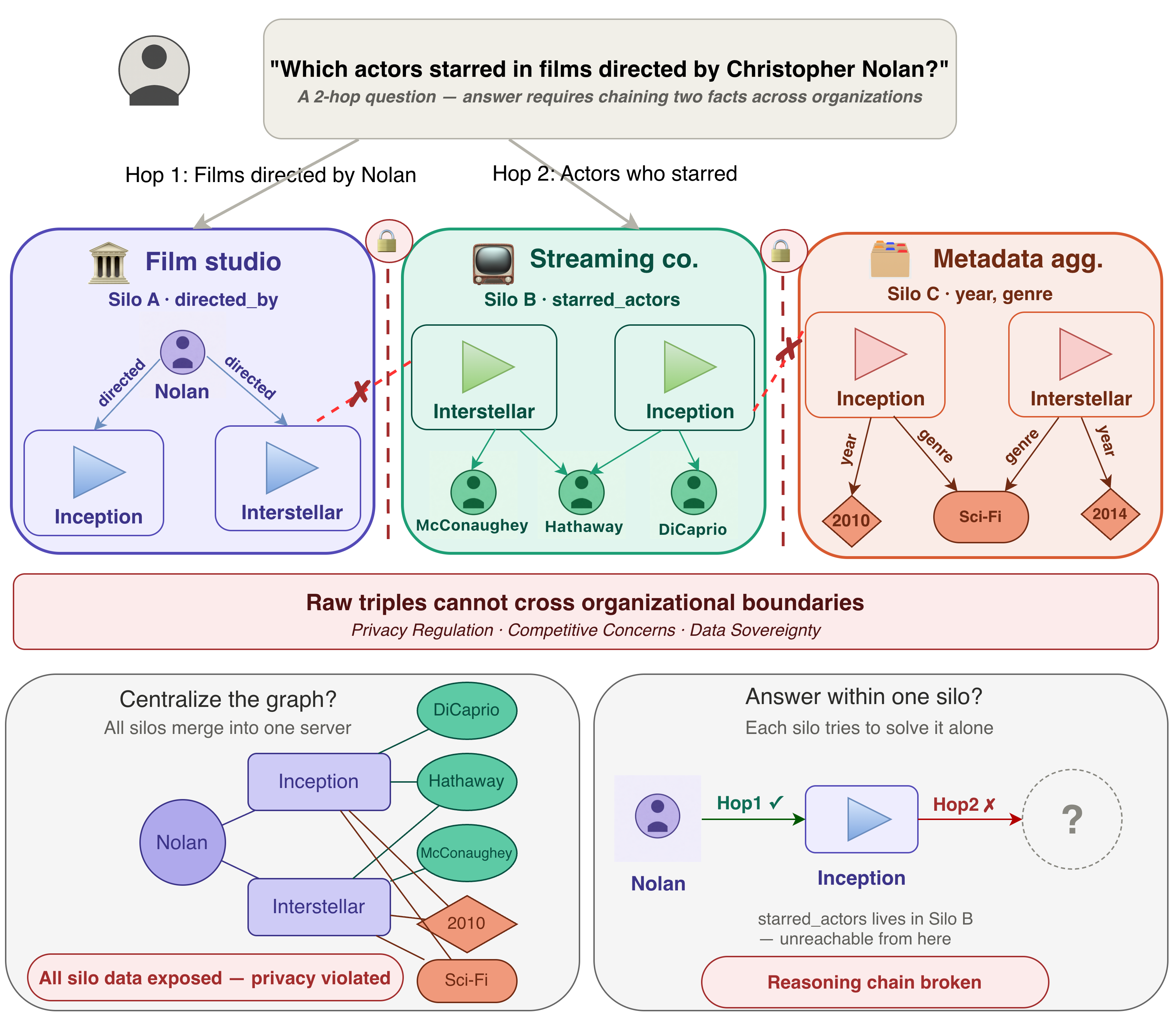}
   \caption{Motivating example of multi-hop KGQA in the vertical federated
setting.}
    \label{fig:fedvkgqa-motivation}
\end{figure}

Answering natural language questions over a knowledge graph (KG) requires chaining facts across multiple entities and relations. While knowledge graph question answering (KGQA) has been studied extensively in the centralized setting where the full graph is accessible to a single system~\cite{chen2024llm,jiang2023unikgqa,luo2024reasoning,ma2025large,pan2024unifying,saxena2020improving,sun2024thinkongraph}, real-world facts are rarely owned by a single organization. For example, a film studio may know a film's director, a streaming platform its actors, and a metadata aggregator its genre. Each organization holds a partial view of the same entities, but sharing raw data across them violates governance constraints, commercial sensitivity, and data sovereignty~\cite{liu2024vertical,ye2025vertical}. 
Consequently, the reasoning chain needed to answer a multi-hop question, which is inherently challenging because it requires connecting two or more facts, is split across organizations by design.
Figure~\ref{fig:fedvkgqa-motivation} makes this concrete. Answering \textit{which
actors starred in films directed by Christopher Nolan} requires two hops across silos. The
first hop resolves the films through \texttt{directed\_by}, which the film studio
owns. The second hop resolves the actors through \texttt{starred\_actors}, which
the streaming platform owns. 
Neither silo can complete the chain independently: centralization exposes private triples, whereas silo-local reasoning breaks the chain.

Federated learning (FL) allows models to be trained over distributed data without sharing raw records across parties~\cite{rahman2023federated}. Most federated knowledge graph embedding (KGE) work studies the horizontal setting, where
different clients hold different triples but share the same relation
vocabulary~\cite{chen2024unaligned,chen2021fede,chen2022federated}. Our problem follows a different structure, where silos share the same entity space but each silo owns a disjoint subset of relation types. This setting corresponds to vertical federated learning (VFL), in which parties share the same sample space while holding different feature views~\cite{khan2024fed,tran2024differentially}. A reasoning path may start in one silo, pass through a shared intermediate entity, and end in another silo. No existing federated method addresses multi-hop reasoning in this structure, and no existing VFL method handles symbolic reasoning over a shared entity space.

We present FedV-KGQA, a framework for multi-hop question answering over vertically partitioned KGs. Each silo trains a local KGE model, while the server concatenates entity embeddings, projects questions into the joint space, and anchors each question to its topic entity. Candidates are ranked by cosine similarity. During QA training, silos receive only their gradient slices; raw triples and relation parameters remain local. This provides structural data separation but no formal privacy guarantee (Section \ref{sec:problem}).

This work addresses three research gaps. First, existing KGQA models
assume centralized graph access and do not handle the case where
relation types are partitioned between
organizations~\cite{liu2025ontology,xu2025harnessing}. Second, prior federated KGE methods target link prediction in the horizontal setting~\cite{sun2026enhancing,zhang2022efficient}, and do not address multi-hop natural language question answering~\cite{hu2025learning,xu2023knowledge}. Third, most existing VFL methods focus on prediction tasks over partitioned feature views and do not address symbolic multi-hop reasoning over a shared entity space~\cite{liu2024vertical,ye2025vertical}. To the best of our
knowledge, FedV-KGQA is the first framework that brings
together vertical federation, multi-hop KGQA, and end-to-end answer
ranking in a unified system. The main contributions of this paper are as follows.
\begin{itemize}
    \item We formulate multi-hop KGQA in a vertical federated setting
    where silos share entity identities but own disjoint relation
    types, a problem setting that, to the best of our knowledge, has
    not been studied for multi-hop KGQA.

    \item We propose FedV-KGQA, a framework that combines
    local KGE training, server-side entity fusion, question projection,
    and silo-specific gradient return for distributed KGQA.

    \item We introduce topic entity anchoring, a mechanism that
    grounds the question vector in the topic entity's fused multi-silo
    embedding, directing the search to the correct graph
    neighborhood without any runtime silo communication.

   \item We evaluate FedV-KGQA across multiple KGE models,
    language encoders, and silo configurations on three benchmarks,
    and show that multi-hop question answering is achievable
    even when the graph is vertically partitioned.
\end{itemize}

\section{Related Work}
\label{sec:related}

Our work spans multi-hop KGQA, federated KGE, question answering over distributed
KGs, and VFL. Table~\ref{tab:related} compares these lines.

\textbf{Multi-Hop KGQA.} Multi-hop KGQA has been studied mainly in the centralized setting, where the
full graph is available during training and inference.
Embedding-based methods such as EmbedKGQA~\cite{saxena2020improving} and
UniKGQA~\cite{jiang2023unikgqa} rank candidate answers by mapping questions
into the KG embedding space. RelChain~\cite{jin2023improving} improves multi-hop KGQA by introducing explicit relational chain reasoning over KG embeddings. 
More recent methods use large language models (LLMs) to guide retrieval and reasoning, including RoG~\cite{luo2024reasoning}, Think-on-Graph~\cite{sun2024thinkongraph}, and GMeLLo~\cite{chen2024llm}, which addresses multi-hop KGQA in evolving environments. These methods support natural language input and multi-hop reasoning, but they
assume centralized graph access or centrally available retrieved evidence.

\textbf{Federated Knowledge Graph Embedding.} Federated KGE methods learn graph representations across distributed clients
while keeping triples local.
FedE~\cite{chen2021fede} is an early framework that aggregates entity embeddings
on a central server.
FedR~\cite{zhang2022efficient} switches to relation embedding aggregation to
reduce susceptibility to embedding inversion attacks.
FKGE~\cite{peng2021differentially} adds differential privacy (DP) noise to shared
embeddings, and FedLU~\cite{zhu2023heterogeneous} addresses heterogeneity and
unlearning. Later work extends this line with alignment and contrastive
objectives~\cite{chen2022federated,chen2024unaligned}.
FedTREK-LM~\cite{spadea2026federated} combines FL, personal KGs, and lightweight language models for decentralized recommendation and KG completion.
All of these methods operate in the horizontal federated setting, where every
client holds triples from the same relation vocabulary but over different subsets
of entities, and they target link prediction rather than natural language question
answering.

\textbf{Question Answering over Distributed Knowledge Graphs.} A smaller line of work considers question answering over distributed KGs.
FedNGDB~\cite{hu2025learning} studies federated neural graph databases for
complex query answering over distributed KGs, supporting multi-hop logical
queries and fusing entity representations from multiple clients.
KG-RL-FL~\cite{xu2023knowledge} explores reinforcement-based federated QA
with KG support in a language-specific setting and accepts natural language
input.
FL-KG-QA~\cite{gunti2025federated} considers federated KGQA with natural
language questions, but focuses on simple question answering rather than
multi-hop reasoning. These works move beyond link prediction, but they do not formulate multi-hop
natural language KGQA in a VFL setting where silos share entity identities but
own disjoint relation types. A separate line answers queries across sources
through federated SPARQL, from FedX~\cite{schwarte2011fedx} and
SPLENDID~\cite{gorlitz2011splendid} to recent engines such as
FedUP~\cite{aimonier2024fedup} and ownership-preserving platforms such as
Pistis~\cite{zhou2025pistis}. These systems require queryable endpoints and
structured queries, whereas our silos expose only embeddings and execute no
query over their triples, so they address a different problem.

\begin{table}[t]
\centering
\setlength{\tabcolsep}{4pt}
\renewcommand{\arraystretch}{0.80}
\caption{System comparison. VFL: vertical split, disjoint relations;
Multi-hop: paths $\geq 2$; KGE fusion: representations fused across silos;
NL: natural language questions.}
\label{tab:related}
\scriptsize
\begin{tabular}{lcccc}
\toprule
System & VFL & Multi-hop & KGE fusion & NL \\
\midrule
EmbedKGQA~\cite{saxena2020improving} & \xmark & \cmark & \xmark & \cmark \\
UniKGQA~\cite{jiang2023unikgqa}      & \xmark & \cmark & \xmark & \cmark \\
RelChain~\cite{jin2023improving}     & \xmark & \cmark & \xmark & \cmark \\
RoG~\cite{luo2024reasoning}          & \xmark & \cmark & \xmark & \cmark \\
GMeLLo~\cite{chen2024llm}            & \xmark & \cmark & \xmark & \cmark \\
\midrule
FedE~\cite{chen2021fede}             & \xmark & \xmark & \cmark & \xmark \\
FedR~\cite{zhang2022efficient}       & \xmark & \xmark & \cmark & \xmark \\
FKGE~\cite{peng2021differentially}   & \xmark & \xmark & \cmark & \xmark \\
FedLU~\cite{zhu2023heterogeneous}    & \xmark & \xmark & \cmark & \xmark \\
\midrule
FedNGDB~\cite{hu2025learning}        & \xmark & \cmark & \cmark & \xmark \\
KG-RL-FL~\cite{xu2023knowledge}      & \xmark & \xmark & \xmark & \cmark \\
FL-KG-QA~\cite{gunti2025federated}   & \xmark & \xmark & \cmark & \cmark \\
\midrule
\textbf{FedV-KGQA (ours)}            & \cmark & \cmark & \cmark & \cmark \\
\bottomrule
\end{tabular}
\end{table}

\textbf{Positioning.} FedV-KGQA differs from each line above along the axes in
Table~\ref{tab:related}. Against centralized multi-hop KGQA, it removes the
assumption of full graph access and ranks answers over embeddings that no single
party can assemble. Against federated KGE, it changes both the split and the
task, since relations are partitioned rather than triples and the objective is
natural language answer ranking rather than link prediction. Against distributed
KGQA, it targets reasoning chains whose hops lie in different silos, which none
of these systems formulate. Against federated SPARQL, it requires no queryable
endpoint and issues no query over silo triples.

\section{Problem Formulation}
\label{sec:problem}

Consider a KG $\mathcal{G} = (\mathcal{E}, \mathcal{R}, \mathcal{T})$,
where $\mathcal{E}$ is the set of entities, $\mathcal{R}$ is the set of relation
types, and $\mathcal{T} \subseteq \mathcal{E} \times \mathcal{R} \times \mathcal{E}$
is the set of relational triples $(h, r, t)$.
In a VFL setting, this graph is partitioned across $K$
independent data silos $\{\mathcal{S}_1, \dots, \mathcal{S}_K\}$ such that each
silo $\mathcal{S}_k$ owns a private relation subset $\mathcal{R}_k \subseteq \mathcal{R}$
satisfying two conditions: the relation sets are pairwise disjoint,
$\mathcal{R}_i \cap \mathcal{R}_j = \emptyset$ for all $i \neq j$, and their
union covers the full relation vocabulary,
$\bigcup_{k=1}^{K} \mathcal{R}_k = \mathcal{R}$. Disjointness is deliberate. It ensures that multi-hop chains
spanning different relation categories must cross a silo boundary. Handling
overlapping relations would require ownership or fusion rules that we do not
address. The entity vocabulary $\mathcal{E}$ is shared and consistently identified across
all participants, while each silo $\mathcal{S}_k$ possesses only a local triple
set $\mathcal{T}_k \subseteq \mathcal{E} \times \mathcal{R}_k \times \mathcal{E}$. We assume a shared entity identifier space exists across all silos prior to federation, following standard VFL practice~\cite{liu2024vertical,ye2025vertical}. Consequently, the complete global triple set $\mathcal{T} = \bigcup_{k=1}^{K}
\mathcal{T}_k$ is never centralized and is never jointly accessible to any
single party.

Given a natural language question $q$ anchored to a topic entity $e_0 \in
\mathcal{E}$, the goal of FedV-KGQA is to predict the correct answer entity
$\hat{e} \in \mathcal{E}$.
The answer is connected to the topic entity through a reasoning path of length
$L \geq 1$; that is, there exist intermediate entities and relations such that
\begin{equation}
  (e_0, r_1, e_1),\;
  (e_1, r_2, e_2),\;
  \dots,\;
  (e_{L-1}, r_L, \hat{e}) \;\in\; \mathcal{T}.
  \label{eq:path}
\end{equation}
While the framework supports any $L \geq 1$, our experiments focus on
multi-hop settings where $L \geq 2$, as these are the cases in which cross-silo
reasoning is required.
The core challenge of this vertical partition is that individual hops in the
reasoning chain may reside in different silos, so the path cannot be resolved
by any single party in isolation.
The learning task is to rank candidate entities for the most plausible answer
while ensuring that, for every silo $\mathcal{S}_k$, the local triples
$\mathcal{T}_k$ and silo-specific relational information, such as the relation
embedding matrix $\mathbf{R}_k$, remain private and are never transmitted to
any other silo or to the central server.

\textbf{Assumptions.} Four assumptions bound the setting we study.
(A1)~Silos share a consistently aligned entity identifier space, established
before federation and treated as given, following standard VFL
practice~\cite{liu2024vertical,ye2025vertical}. (A2)~Each question is provided
together with its topic entity $e_0$, as supplied by the benchmarks.
(A3)~Silos agree in advance on a schema-level rule set $\mathcal{O}$ that
contains relation axioms only, with no instances and no triples.
(A4)~The graph is static, and candidate sets are precomputed offline.

\section{Methodology}
\label{sec:method}

\begin{figure}[t]
    \centering
    \includegraphics[
        width=\linewidth,
        height=0.45\textheight,
        keepaspectratio
    ]{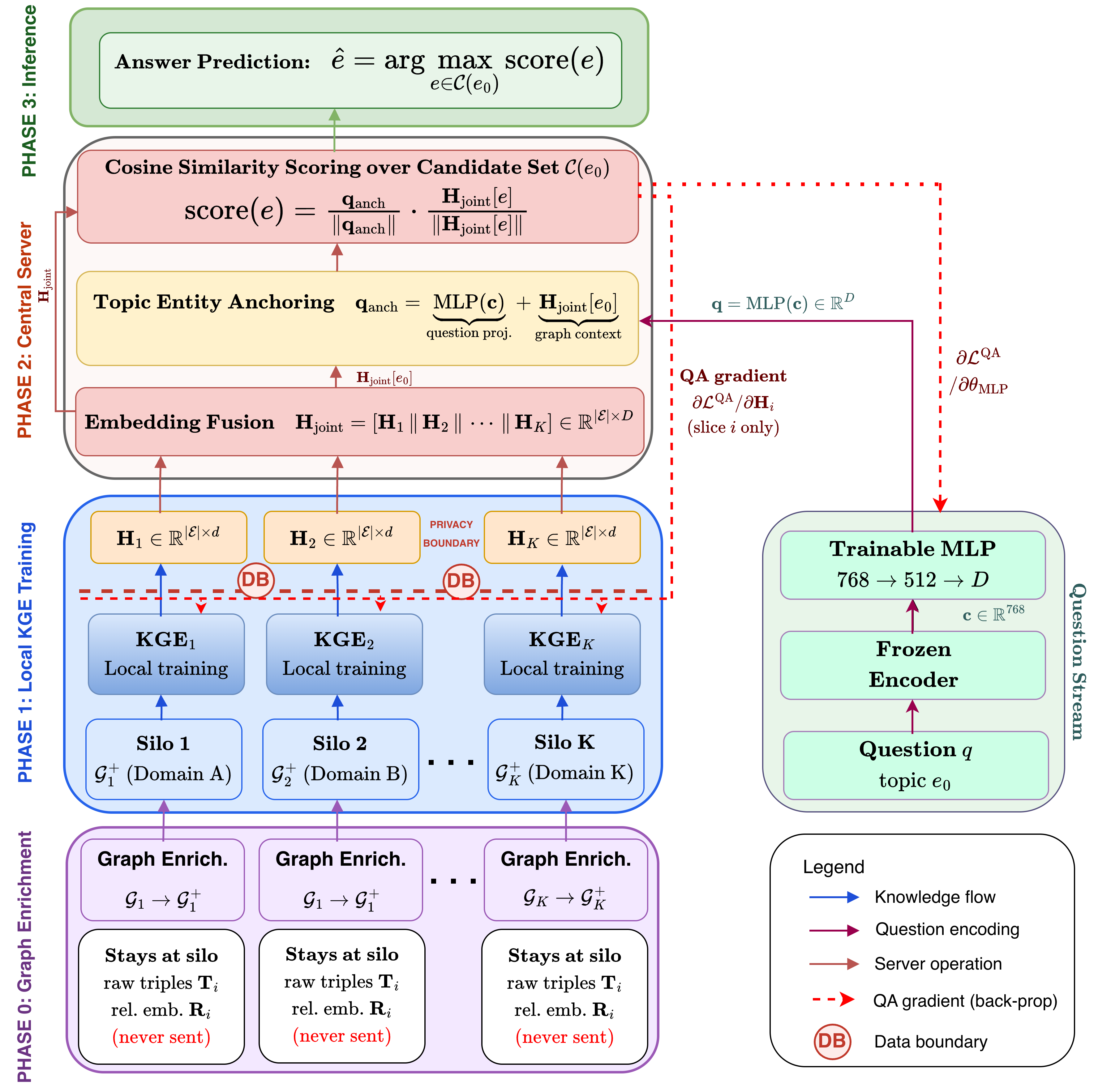}
    \caption{Overview of the FedV-KGQA architecture.}
    \label{fig:fedvkgqa-diagram}
\end{figure}

FedV-KGQA operates in four phases, as shown in Figure~\ref{fig:fedvkgqa-diagram}: local graph enrichment, local KGE training, server-side fusion and QA training, and inference.

\subsection{Phase 0: Local Graph Enrichment}
\label{sec:phase0}
Before local training starts, all silos agree on a shared T-Box
$\mathcal{O}$ that contains only relation rules. We apply local graph enrichment within each silo to improve entity  representation quality in Phase~1 and ensure answer entities are 
reachable within the candidate set. More specifically, each
silo $\mathcal{S}_k$ applies these shared rules to its local triple set
$\mathcal{T}_k$ and thereby produces an enriched set $\hat{\mathcal{T}}_k$.

First, inverse property axioms add the triple $(t, r^{-}, h)$ for every
$(h, r, t) \in \mathcal{T}_k$ with a declared inverse relation $r^{-}$. This matters because answer entities often appear only as tails in their
originating silo. In a silo holding \textit{starred\_actor} triples, an actor
never appears as a head, so its embedding is trained from one role only and carries a weaker signal for cosine scoring. The inverse triple restores the head role and strengthens the representation. Second, property chain axioms add $(h, r_{\text{chain}}, t)$ for every pair
$(h, r_1, m)$ and $(m, r_2, t)$ in $\mathcal{T}_k$ whenever $\mathcal{O}$
declares $r_1 \circ r_2 \rightarrow r_{\text{chain}}$. Here, both $r_1$ and
$r_2$ belong to the same silo, since each silo applies chain axioms only
within its own relation subset. In this way, chain axioms create a direct
one-hop triple between the topic entity and a semantically related entity
inside the same silo. Therefore, they reduce the need to infer missing
relational links during answer search. Throughout this process, no triple leaves the silo. $\mathcal{O}$ declares only
inverse and two-relation chain axioms, and our 3-hop experiments reuse this same
rule set, extending only the bounded candidate expansion.

In addition, we construct candidate sets in Phase~0 so that they can be reused
during both training and inference without repeated graph traversal. For each topic entity $e_0$, we perform a two-hop expansion in which silos first
contribute their one-hop neighbors, which are combined before the second hop:

\begin{align}
\mathcal{N}_1(e_0) &= \bigcup_k \bigl\{e' : \exists\, r,\;
   (e_0, r, e') \in \hat{\mathcal{T}}_k \;\text{or}\;
   (e', r, e_0) \in \hat{\mathcal{T}}_k\bigr\}, \\
\mathcal{C}(e_0) &= \mathcal{N}_1(e_0) \cup \bigcup_k \bigcup_{m \in \mathcal{N}_1(e_0)}
   \mathcal{N}_1^{k}(m).
\end{align}
   
Thus, $\mathcal{N}_{1}(e_0)$ contains the one-hop neighbors of the topic entity
across all silos, whereas $\mathcal{C}(e_0)$ collects entities that become
reachable within two hops. Because the first hop is combined before the second
expands, a chain whose hops lie in different silos remains reachable, which is
the case that cross-silo reasoning requires. Moreover, for the relation patterns
defined in $\mathcal{O}$, local graph enrichment further improves reachability of
the gold answer. In a privacy-preserving deployment, the union in each expansion
can be computed through a Private Set Union
protocol~\cite{dong2025efficient,tu2025fast}, which reveals only the final union
and does not disclose which silo contributed which candidates.

\subsection{Phase 1: Local Knowledge Graph Embedding}
\label{sec:phase1}

Next, in Phase~1, each silo $\mathcal{S}_k$ independently trains a KGE model
on its enriched triple set $\hat{\mathcal{T}}_k$ without communicating with any
other party. In this way, every silo learns its own local structural view of
the KG while keeping its triples and relation parameters private. More
specifically, the model learns an entity embedding matrix
$\mathbf{H}_k \in \mathbb{R}^{|\mathcal{E}| \times d_k}$ and a relation
embedding matrix $\mathbf{R}_k$. To study the effect of different embedding
methods, we evaluate four KGE scoring functions, which are summarized in
Table~\ref{tab:kge}. All models are trained with a margin ranking loss:
\begin{equation}
  \mathcal{L}^{\text{KGE}}_k
  = \sum_{(h,r,t)\,\in\,\hat{\mathcal{T}}_k}
    \max\!\bigl(0,\;
      \gamma + \phi(h, r, t^{-}) - \phi(h, r, t)
    \bigr),
  \label{eq:kge-loss}
\end{equation}

where $t^{-}$ is a randomly sampled negative tail entity and $\gamma$ is the
margin. The objective scores a true triple above a corrupted one by at least $\gamma$,
so the learned embeddings capture local relational structure. At the end of Phase~1, the relation matrix $\mathbf{R}_k$ and the enriched triples $\hat{\mathcal{T}}_k$
remain entirely inside silo $\mathcal{S}_k$, whereas only the entity embedding
matrix $\mathbf{H}_k$ is transmitted to the server. This embedding-only transmission serves as the data boundary shown in Figure~\ref{fig:fedvkgqa-diagram}, since raw triples and relation parameters never leave the silo.

\begin{table}[t]
\centering
\setlength{\tabcolsep}{4pt}
\renewcommand{\arraystretch}{0.90}
\caption{KGE scoring functions ($d_k = d$: real-valued; $d_k = 2d$: complex-valued).}
\label{tab:kge}
\scriptsize
\begin{tabular}{llcc}
\toprule
Model & Scoring function $\phi(h,r,t)$ & Space & $d_k$ \\
\midrule
TransE~\cite{bordes2013translating}
  & $-\|\mathbf{h} + \mathbf{r} - \mathbf{t}\|_2$
  & $\mathbb{R}^d$ & $d$ \\
DistMult~\cite{yang2014embedding}
  & $\sum_j h_j r_j t_j$
  & $\mathbb{R}^d$ & $d$ \\
RotatE~\cite{sun2019rotate}
  & $-\|\mathbf{h} \circ e^{i\mathbf{r}} - \mathbf{t}\|_2$
  & $\mathbb{C}^d$ & $2d$ \\
ComplEx~\cite{trouillon2016complex}
  & $\operatorname{Re}(\langle \mathbf{h}, \mathbf{r},
    \bar{\mathbf{t}} \rangle)$
  & $\mathbb{C}^d$ & $2d$ \\
\bottomrule

\end{tabular}
\end{table}

\subsection{Phase 2: Server-Side Fusion and QA Training}
\label{sec:phase2}
After KGE training is completed, the server receives the
embeddings from all silos and combines them to support answer ranking. More
specifically, it concatenates the per-silo embeddings to form a joint
representation for each entity:
\begin{equation}
  \mathbf{H}_{\text{joint}}
  = \bigl[\mathbf{H}_1 \;\|\; \mathbf{H}_2 \;\|\; \cdots \;\|\;
    \mathbf{H}_K\bigr]
  \in \mathbb{R}^{|\mathcal{E}| \times D},
  \quad D = \textstyle\sum_{k=1}^K d_k.
  \label{eq:fusion}
\end{equation}
In this way, the server preserves each silo's geometric view in separate
dimensions. By contrast, sum or mean fusion would mix embeddings learned in
different geometric spaces and would weaken the silo-specific
signal~\cite{khan2024fed}. The concatenation order is fixed at setup and shared by all parties. Since the
multilayer perceptron (MLP) learns a projection into this layout, the model is not invariant to silo
permutations, and adding or removing a silo changes $D$ and requires retraining
the projection head. 

Next, we encode the question $q$ using a frozen pre-trained transformer,
namely BERT~\cite{devlin2019bert}, DistilBERT~\cite{sanh2019distilbert}, or
RoBERTa~\cite{liu2019roberta}, to obtain a contextual representation
$\mathbf{c} = \operatorname{Enc}(q)[\texttt{CLS}] \in \mathbb{R}^{768}$.
We then use a trainable two-layer MLP to project $\mathbf{c}$ into the joint
embedding space, that is,
$\mathbf{q} = \operatorname{MLP}(\mathbf{c}) \in \mathbb{R}^D$.
Here, we freeze the encoder so that only the MLP parameters
$\theta_{\text{MLP}}$ are updated on the server. As a result, the linguistic
knowledge learned during pre-training is preserved. However, the projected vector $\mathbf{q}$ only captures the expected answer
type and does not indicate where in the graph the search should begin.
Therefore, we anchor the question to the topic entity $e_0$ by adding its
joint embedding:
\begin{equation}
  \mathbf{q}_{\text{anch}}
  = \mathbf{q} + \mathbf{H}_{\text{joint}}[e_0].
  \label{eq:anchoring}
\end{equation}
Through this step, the question representation is grounded in the
neighborhood of $e_0$, which directs the search toward the relevant part of
the graph. Importantly, this requires no runtime communication with any silo,
since $\mathbf{H}_{\text{joint}}$ is already available on the server. We then rank each candidate $e \in \mathcal{C}(e_0)$ by cosine similarity:
\begin{equation}
  \mathrm{score}(e)
  = \frac{\mathbf{q}_{\text{anch}}^{\top}
          \mathbf{H}_{\text{joint}}[e]}
         {\|\mathbf{q}_{\text{anch}}\|\;
          \|\mathbf{H}_{\text{joint}}[e]\|}.
  \label{eq:scoring}
\end{equation}
Based on these scores, QA training minimizes a margin ranking loss over the
candidate set:
\begin{equation}
  \mathcal{L}^{\text{QA}}
  = \max\!\left(0,\;
    \gamma
    + \max_{e^{-} \in \mathcal{C}(e_0) \setminus \mathcal{A}}
        \mathrm{score}(e^{-})
    - \max_{e^{+} \in \mathcal{A}}
        \mathrm{score}(e^{+})
    \right),
  \label{eq:qa-loss}
\end{equation}
where $\mathcal{A}$ is the set of gold answer entities. This objective 
penalizes the model when the hardest negative candidate scores too close 
to the best gold answer. Finally, because $\mathbf{H}_{\text{joint}}$ is 
formed by concatenation, the gradient $\partial\mathcal{L}^{\text{QA}} / 
\partial\mathbf{H}_{\text{joint}}$ naturally decomposes along the 
concatenation dimension. Thus, the slice $\partial\mathcal{L}^{\text{QA}} 
/ \partial\mathbf{H}_k$ occupies a contiguous block of columns that 
corresponds only to silo $k$. The server returns this slice to silo $k$, 
and the silo uses it to update $\mathbf{H}_k$ locally. Meanwhile, relation 
embeddings $\mathbf{R}_k$ receive no gradient update and remain frozen at 
their Phase~1 values. In this way, no information about any other silo's 
embeddings or triples is exposed during QA training. This upload-download 
exchange constitutes one communication round per QA training epoch, and 
the total number of rounds equals the number of training epochs $T$. The boundary this establishes is structural rather than formal. The only
quantity leaving each silo is $\mathbf{H}_k$, but this does not bound what an
honest-but-curious server could infer from the embeddings it receives, and we
make no formal DP claim.

\subsection{Phase 3: Inference}
\label{sec:phase3}

At inference time, each silo transmits its fine-tuned $\mathbf{H}_k$ to the
server. The server then forms $\mathbf{H}_{\text{joint}}$, retrieves the
precomputed candidate set $\mathcal{C}(e_0)$ for the topic entity, encodes the
question, and computes the anchored question vector $\mathbf{q}_{\text{anch}}$
using Equation~\eqref{eq:anchoring}. Next, it scores every candidate in
$\mathcal{C}(e_0)$ with Equation~\eqref{eq:scoring} and returns the entity
with the highest score:
\begin{equation}
  \hat{e}
  = \arg\max_{e \in \mathcal{C}(e_0)}
    \mathrm{score}(e).
  \label{eq:inference}
\end{equation}
Therefore, the full inference pipeline reduces to matrix lookups, one MLP
forward pass, one anchoring step, and dot products over the candidate set.
Moreover, no gradient is computed at this stage, and no inter-silo
communication occurs beyond providing $\mathbf{H}_k$. 

\textbf{Complexity and Space.} Graph enrichment and candidate construction are
one-time offline steps. Inverse axioms add at most $|\mathcal{T}_k|$ triples,
while chain axioms dominate at $\mathcal{O}(\sum_{m} d^{-}(m)\, d^{+}(m))$ over
intermediate entities $m$, where $d^{-}$ and $d^{+}$ are in-degree and
out-degree. In Phase~1, training in each silo costs
$\mathcal{O}(|\hat{\mathcal{T}}_k| d)$ per epoch in parallel and stores
$\mathcal{O}((|\mathcal{E}| + |\mathcal{R}_k|) d)$ parameters, so cost is
bounded by the largest silo. Candidate scoring in Phase~2 costs
$\mathcal{O}(|\mathcal{C}(e_0)| K d)$ per iteration, which is cheap since
$|\mathcal{C}(e_0)| \ll |\mathcal{E}|$. The server holds
$\mathbf{H}_{\text{joint}}$ at $\mathcal{O}(|\mathcal{E}| K d)$, the dominant
space term, and communication per round is $\mathcal{O}(|\mathcal{E}| d)$ per
silo, independent of local triple count.

\section{Experiments}
We evaluate FedV-KGQA through six research questions. \textbf{(RQ1)}~How does it
perform across KGE models and question encoders? \textbf{(RQ2)}~How well does it
generalize from 2-hop to 3-hop reasoning? \textbf{(RQ3)}~What does each component
contribute? \textbf{(RQ4)}~How does it compare against adapted baselines?
\textbf{(RQ5)}~How robust is it to embedding perturbation? \textbf{(RQ6)}~What
communication cost is required to reach a target performance level?

\subsection{Datasets}

We conduct experiments on three widely used KGQA benchmarks that differ in
domain, scale, and question complexity.

\textbf{MetaQA}~\cite{zhang2018variational} is a movie-domain benchmark built
from the WikiMovies knowledge base. It contains more than 43{,}000 entities and
9 relation types covering movie metadata such as directors, actors, genres, and
release years. The dataset provides questions at 1-hop, 2-hop, and 3-hop
depths. We use the 2-hop split as our main evaluation setting and extend to the
3-hop split to study multi-hop generalization.

\textbf{PathQuestion (PQ-2H/PQ-3H)}~\cite{zhou2018interpretable} is constructed
from a subset of Freebase13 and focuses on person-centric relations such as
family ties, demographic attributes, and biographical facts. PQ-2H contains
2-hop questions and PQ-3H contains 3-hop questions. Its relatively small size
makes it a challenging benchmark for learning robust multi-hop reasoning
patterns.

\textbf{WebQuestionsSP (WebQSP)}~\cite{yih2016value} is derived from the
original WebQuestions benchmark and grounded in the full Freebase KG. It covers diverse domains such as people, places, organizations, and
entertainment, with over 985{,}000 entities and up to 2-hop reasoning
questions. WebQSP provides the most challenging setting in our evaluation due
to the scale of its KG and the open-domain nature of its
questions.

\subsection{Experimental Setup}
\label{sec:5.2}
\textbf{Evaluation Metrics.}
We use two ranking metrics. Mean Reciprocal Rank (MRR) measures how highly
the first correct answer is ranked on average:
$\text{MRR} = \frac{1}{|Q|} \sum_{i=1}^{|Q|} \frac{1}{\text{rank}_i}$.
Hits@K (H@K) measures the fraction of questions with a correct answer in
the top $K$ positions. We report MRR, H@3, H@5, and H@10.

\textbf{Model Configurations.}
We evaluate 12 model configurations.
We test four KGE models (TransE, DistMult, ComplEx, RotatE) paired with
three frozen encoders (BERT, DistilBERT, RoBERTa), giving 12 combinations
per dataset. TransE and DistMult use $d{=}256$ real-valued embeddings.
ComplEx and RotatE use $2d{=}512$ real-valued dimensions per silo.
The joint embedding dimension is $Kd$ or $K \times 2d$ for $K$ silos.
Each encoder outputs a vector of dimension 768, projected by a two-layer
MLP into the joint space. Only the MLP head is updated during QA training.

\textbf{Training and Candidate Filtering.}
All experiments use NVIDIA H100 NVL GPUs (96\,GB) with CUDA 12.2. In the KGE phase, each silo trains for 100 epochs using
Adam with learning rate $10^{-3}$, margin $\gamma{=}1.0$, batch size 512, and
10 negative samples per triple. In the QA phase, we use Adam with learning rate
$10^{-4}$, batch size 64, and margin $\gamma{=}1.0$ for 100 epochs. We select
the best checkpoint on the development set. Gradient norms are clipped to 1.0
for the MLP and entity embeddings. For candidate filtering, we
expand two hops from the topic entity. On MetaQA and PathQuestion, one-hop
and two-hop neighbors are capped at 50 and 20
($\text{max\_neighbors}{=}100$). On WebQSP, we increase these caps to 100, 30,
and $\text{max\_neighbors}{=}200$ to accommodate its denser graph with more
than 985{,}000 entities. For 3-hop experiments, we add a third hop capped at
10 neighbors per node.

\textbf{Silo Configurations.}
No public benchmark provides naturally vertical multi-hop KGQA data, so we
partition each dataset into 3, 5, and 7 silos by semantic relation category. For example, in MetaQA with three silos, Silo~A holds directorial and
writing relations, Silo~B holds cast and tag relations, and Silo~C holds genre,
language, release year, and ratings. Each relation belongs to exactly one silo,
and entities are shared across all silos. In the three-silo configuration, and counting inverse relations, the MetaQA silos contain 4, 3, and 8 relations over 63K, 129K, and 84K enriched triples, respectively; the PathQuestion silos contain 3, 5, and 5 relations over 18K, 168K, and 191K triples, respectively; and the WebQSP silos contain 332, 847, and 2,098 relations over 0.59M, 0.99M, and 1.89M triples, respectively.
Partitions therefore differ
substantially in how much structure each silo can learn.
Under the Silo-3 partition, the combined candidate set $\mathcal{C}(e_0)$
contains a gold answer for 99\% of MetaQA questions, 100\% on PathQuestion, and
78\% on WebQSP. The same set, when built from any one silo alone, achieves at most 54\%, 35\%, and 46\%, respectively, which confirms that no partition holds a complete reasoning chain. 
Recall bounds attainable accuracy, since answers outside $\mathcal{C}(e_0)$
cannot be ranked. WebQSP therefore starts from a lower ceiling than the other
two benchmarks.

\subsection{RQ1: Performance on 2-Hop Reasoning}
Tables~\ref{tab:MetaQA},~\ref{tab:CWQ}, and~\ref{tab:pathquestion} present
the full results across all 12 model configurations and three silo
partitions. We highlight the key findings below.

\textbf{KGE model comparison.}
On MetaQA, all four KGE models perform similarly, with MRR ranging from
0.71 to 0.76 on the 3-silo partition. The movie-domain graph has simple
relational patterns, so all models capture it well. On WebQSP, TransE
clearly outperforms the others. With BERT on Silo-3, TransE reaches an MRR
of 0.54 while DistMult achieves only 0.41. Part of WebQSP's gap comes from candidate coverage, since only
78\% of its questions have a gold answer in the candidate set. On PathQuestion, TransE again
performs strongly, though DistMult and RotatE are competitive in some
settings. Overall, TransE is the most stable model across all three
benchmarks. Complex-valued models show no consistent advantage despite using $2d{=}512$ real dimensions per silo, twice the budget of TransE and DistMult.

\begin{table}[t]
\centering
\caption{Results on the MetaQA dataset. The bold font denotes the best result.}
\label{tab:MetaQA}
\renewcommand{\arraystretch}{0.96}
\scriptsize
\resizebox{\textwidth}{!}{%
\begin{tabular}{ll|cccc|cccc|cccc}
\hline
KGE & Encoder & \multicolumn{4}{c|}{Silo-3 } & \multicolumn{4}{c|}{Silo-5} & \multicolumn{4}{c}{Silo-7} \\
 & & MRR & H@3 & H@5 & H@10 & MRR & H@3 & H@5 & H@10 & MRR & H@3 & H@5 & H@10 \\
\hline
\multirow{3}{*}{TransE} 
 & BERT       & \textbf{0.76} & \textbf{0.83} & 0.88 & 0.93 & \textbf{0.83} & \textbf{0.87} & \textbf{0.89} & 0.91 & \textbf{0.74} & 0.78 & 0.80 & \textbf{0.84} \\
 & DistilBERT & \textbf{0.76} & \textbf{0.83} & \textbf{0.89} & \textbf{0.94} & 0.82 & 0.86 & 0.88 & 0.91 & \textbf{0.74} & 0.79 & \textbf{0.81} & 0.82 \\
 & RoBERTa    & 0.75 & 0.82 & 0.88 & 0.93 & 0.79 & 0.85 & 0.88 & \textbf{0.92} & 0.58 & 0.71 & 0.79 & 0.81 \\
\hline
\multirow{3}{*}{DistMult} 
 & BERT       & \textbf{0.76} & 0.82 & 0.87 & 0.93 & \textbf{0.83} & 0.86 & \textbf{0.89} & \textbf{0.92} & 0.69 & 0.77 & 0.79 & 0.82 \\
 & DistilBERT & \textbf{0.76} & 0.82 & 0.88 & 0.93 &  \textbf{0.83} & \textbf{0.87} & \textbf{0.89} & \textbf{0.92} & 0.70 & 0.77 & 0.79 & 0.82 \\
 & RoBERTa    & 0.73 & 0.80 & 0.86 & 0.92 & 0.79 & 0.85 & \textbf{0.89} & \textbf{0.92} & 0.58 & 0.70 & 0.78 & 0.81 \\
\hline
\multirow{3}{*}{ComplEx} 
 & BERT       & 0.74 & 0.80 & 0.86 & 0.91 & \textbf{0.83} & \textbf{0.87} & \textbf{0.89} & \textbf{0.92} & 0.71 & 0.78 & 0.80 & 0.82 \\
 & DistilBERT & \textbf{0.76} & 0.81 & 0.87 & 0.92 & \textbf{0.83} & \textbf{0.87} & \textbf{0.89} & \textbf{0.92} & \textbf{0.74} & \textbf{0.80} & \textbf{0.81} & 0.82 \\
 & RoBERTa    & 0.72 & 0.80 & 0.86 & 0.91 & 0.80 & 0.85 & 0.88 & \textbf{0.92} & 0.59 & 0.70 & 0.78 & 0.80 \\
\hline
\multirow{3}{*}{RotatE} 
 & BERT       & 0.73 & 0.80 & 0.85 & 0.91 & 0.82 & 0.86 & 0.88 & 0.91 & 0.70 & 0.79 & \textbf{0.81} & 0.83 \\
 & DistilBERT & 0.73 & 0.80 & 0.86 & 0.91 & 0.82 & 0.86 & \textbf{0.89} & 0.91 & 0.71 & 0.79 & \textbf{0.81} & 0.83 \\
 & RoBERTa    & 0.71 & 0.79 & 0.85 & 0.91 & 0.79 & 0.84 & 0.88 & 0.91 & 0.60 & 0.72 & 0.78 & 0.81 \\
\hline
\end{tabular}}
\end{table}

\begin{table}[t]
\centering
\caption{Results on the WebQSP dataset. The bold font denotes the best result.}
\label{tab:CWQ}
\renewcommand{\arraystretch}{0.96}
\scriptsize
\resizebox{\textwidth}{!}{%
\begin{tabular}{ll|cccc|cccc|cccc}
\hline
KGE & Encoder & \multicolumn{4}{c|}{Silo-3 } & \multicolumn{4}{c|}{Silo-5} & \multicolumn{4}{c}{Silo-7} \\
 & & MRR & H@3 & H@5 & H@10 & MRR & H@3 & H@5 & H@10 & MRR & H@3 & H@5 & H@10 \\
\hline
\multirow{3}{*}{TransE} 
 & BERT       & \textbf{0.54} & \textbf{0.59} & \textbf{0.64} & 0.71 & \textbf{0.52} & \textbf{0.57} & 0.62 & 0.69 & \textbf{0.51} & \textbf{0.56} & \textbf{0.62} & 0.67 \\
 & DistilBERT & 0.53 & \textbf{0.59} & \textbf{0.64} & \textbf{0.73} & \textbf{0.52} & \textbf{0.57} & \textbf{0.64} & \textbf{0.71} & 0.50 & \textbf{0.56} & \textbf{0.62} & \textbf{0.69} \\
 & RoBERTa    & 0.43 & 0.49 & 0.54 & 0.61 & 0.41 & 0.47 & 0.53 & 0.60 & 0.39 & 0.44 & 0.50 & 0.57 \\
\hline
\multirow{3}{*}{DistMult} 
 & BERT       & 0.41 & 0.45 & 0.50 & 0.58 & 0.42 & 0.46 & 0.51 & 0.58 & 0.43 & 0.47 & 0.54 & 0.60 \\
 & DistilBERT & 0.40 & 0.43 & 0.49 & 0.57 & 0.41 & 0.45 & 0.50 & 0.57 & 0.41 & 0.45 & 0.51 & 0.57 \\
 & RoBERTa    & 0.27 & 0.28 & 0.31 & 0.38 & 0.27 & 0.28 & 0.31 & 0.39 & 0.27 & 0.28 & 0.32 & 0.37 \\
\hline
\multirow{3}{*}{ComplEx} 
 & BERT       & 0.48 & 0.53 & 0.60 & 0.67 & 0.48 & 0.53 & 0.60 & 0.67 & 0.45 & 0.50 & 0.56 & 0.63 \\
 & DistilBERT & 0.48 & 0.53 & 0.59 & 0.66 & 0.47 & 0.52 & 0.59 & 0.67 & 0.45 & 0.51 & 0.58 & 0.65 \\
 & RoBERTa    & 0.36 & 0.41 & 0.46 & 0.53 & 0.32 & 0.34 & 0.39 & 0.45 & 0.30 & 0.31 & 0.35 & 0.41 \\
\hline
\multirow{3}{*}{RotatE} 
 & BERT       & 0.49 & 0.54 & 0.60 & 0.66 & 0.51 & 0.56 & 0.62 & 0.68 & 0.49 & 0.55 & 0.61 & 0.68 \\
 & DistilBERT & 0.49 & 0.53 & 0.60 & 0.67 & 0.51 & \textbf{0.57} & 0.62 & 0.69 & 0.50 & 0.55 & 0.61 & 0.68 \\
 & RoBERTa    & 0.38 & 0.40 & 0.46 & 0.55 & 0.37 & 0.40 & 0.46 & 0.54 & 0.32 & 0.35 & 0.40 & 0.47 \\
\hline
\end{tabular}%
}
\end{table}

\begin{table}[t]
\centering
\caption{Results on the PathQuestion dataset. The bold font denotes the best result.}
\label{tab:pathquestion}
\renewcommand{\arraystretch}{0.96}
\scriptsize
\resizebox{\textwidth}{!}{%
\begin{tabular}{ll|cccc|cccc|cccc}
\hline
KGE & Encoder & \multicolumn{4}{c|}{Silo-3 } & \multicolumn{4}{c|}{Silo-5} & \multicolumn{4}{c}{Silo-7} \\
 & & MRR & H@3 & H@5 & H@10 & MRR & H@3 & H@5 & H@10 & MRR & H@3 & H@5 & H@10 \\
\hline
\multirow{3}{*}{TransE} 
 & BERT       & 0.65 & \textbf{0.81} & \textbf{0.89} & \textbf{0.96} & \textbf{0.68} & 0.84 & \textbf{0.90} & \textbf{0.96} & 0.55 & 0.71 & 0.78 & 0.90 \\
 & DistilBERT & 0.64 & 0.76 & 0.87 & 0.93 & 0.65 & 0.79 & 0.86 & 0.94 & 0.63 & 0.76 & 0.81 & 0.93 \\
 & RoBERTa    & 0.44 & 0.53 & 0.75 & 0.87 & 0.50 & 0.67 & 0.81 & 0.91 & 0.48 & 0.64 & 0.75 & 0.91 \\
\hline
\multirow{3}{*}{DistMult} 
 & BERT       & 0.64 & 0.80 & 0.85 & 0.93 & 0.63 & 0.81 & 0.88 & 0.94 & 0.62 & \textbf{0.79} & \textbf{0.87} & 0.91 \\
 & DistilBERT & \textbf{0.66} & 0.79 & 0.88 & 0.94 & 0.63 & 0.76 & 0.87 & 0.92 & 0.63 & 0.77 & 0.84 & 0.93 \\
 & RoBERTa    & 0.46 & 0.61 & 0.73 & 0.88 & 0.46 & 0.59 & 0.73 & 0.83 & 0.48 & 0.64 & 0.80 & 0.89 \\
\hline
\multirow{3}{*}{ComplEx} 
 & BERT       & 0.64 & 0.79 & 0.81 & 0.89 & 0.60 & 0.78 & 0.78 & 0.86 & 0.60 & \textbf{0.79} & 0.81 & 0.87 \\
 & DistilBERT & 0.63 & 0.77 & 0.84 & 0.87 & 0.61 & 0.78 & 0.83 & 0.86 & 0.61 & 0.76 & 0.81 & 0.91 \\
 & RoBERTa    & 0.45 & 0.55 & 0.65 & 0.81 & 0.45 & 0.62 & 0.69 & 0.84 & 0.45 & 0.57 & 0.68 & 0.85 \\
\hline
\multirow{3}{*}{RotatE} 
 & BERT       & 0.55 & 0.76 & 0.78 & 0.85 & 0.66 & \textbf{0.85} & 0.89 & 0.95 & 0.63 & 0.77 & \textbf{0.87} & 0.94 \\
 & DistilBERT & 0.57 & 0.73 & 0.77 & 0.83 & 0.64 & 0.79 & \textbf{0.90} & 0.94 & \textbf{0.66} & \textbf{0.79} & 0.86 & \textbf{0.95} \\
 & RoBERTa    & 0.46 & 0.61 & 0.74 & 0.83 & 0.46 & 0.61 & 0.77 & 0.92 & 0.45 & 0.60 & 0.75 & 0.89 \\
\hline
\end{tabular}%
}
\end{table}

\begin{table}[t]
\centering
\caption{Comparative 2-hop and 3-hop performance on the 3-silo partition.}
\label{tab:hop-comparison-full}
\renewcommand{\arraystretch}{0.96}
\scriptsize
\resizebox{\textwidth}{!}{%
\begin{tabular}{ll|cc|cc|cc|cc}
\hline
\textbf{Dataset} & \textbf{Encoder + KGE} & \multicolumn{2}{c|}{\textbf{MRR}} & \multicolumn{2}{c|}{\textbf{H@3}} & \multicolumn{2}{c|}{\textbf{H@5}} & \multicolumn{2}{c}{\textbf{H@10}} \\
 & & 2-Hop & 3-Hop & 2-Hop & 3-Hop & 2-Hop & 3-Hop & 2-Hop & 3-Hop \\
\hline
\multirow{3}{*}{MetaQA}
 & BERT + TransE        & 0.76 & 0.74 & 0.83 & 0.82 & 0.88 & 0.87 & 0.93 & 0.93 \\
 & DistilBERT + ComplEx & 0.76 & 0.72 & 0.81 & 0.80 & 0.87 & 0.86 & 0.92 & 0.91 \\
 & RoBERTa + TransE     & 0.75 & 0.73 & 0.82 & 0.81 & 0.88 & 0.87 & 0.93 & 0.92 \\
\hline
\multirow{3}{*}{PathQuestion}
 & BERT + TransE        & 0.65 & 0.57 & 0.81 & 0.69 & 0.89 & 0.80 & 0.96 & 0.90 \\
 & DistilBERT + DistMult & 0.66 & 0.51 & 0.79 & 0.61 & 0.88 & 0.69 & 0.94 & 0.81 \\
 & RoBERTa + TransE     & 0.44 & 0.37 & 0.53 & 0.43 & 0.75 & 0.53 & 0.87 & 0.67 \\
\hline
\end{tabular}}
\end{table}

\textbf{Encoder comparison.}
BERT and DistilBERT perform similarly across three datasets.
DistilBERT is noteworthy: it has only 66M parameters versus BERT's 110M,
yet matches BERT's QA accuracy. Since the encoder is frozen and only the
MLP head is trained, DistilBERT offers an efficiency advantage.
RoBERTa underperforms both encoders. On MetaQA Silo-7, its
MRR is 0.58 with TransE, compared to 0.74 for BERT and DistilBERT. On
WebQSP Silo-3, it reaches 0.43 versus 0.54 for BERT. The gap suggests
that RoBERTa's \texttt{[CLS]} representations are less compatible with
KGE embedding spaces.

\textbf{Effect of silo count.}
Performance is not monotonic in the silo count. On MetaQA, MRR
falls only from 0.76 to 0.74 for BERT+TransE, and Silo-5 outperforms Silo-3 in
most settings. On WebQSP, the drop is also small, from 0.54 to 0.51. PathQuestion shows a mixed pattern, and some configurations such
as RotatE+DistilBERT improve with more silos. Silo count alone therefore does
not predict performance.

\subsection{ RQ2: Extension to 3-Hop Reasoning}

Table~\ref{tab:hop-comparison-full} compares 2-hop and 3-hop performance on
MetaQA and PathQuestion using the Silo-3 partition. For each encoder, we
select the best-performing KGE model from the 2-hop results and extend it to
the 3-hop setting. On MetaQA, the transition from 2-hop to 3-hop causes only
a minor drop: BERT+TransE decreases by 0.02 in MRR, from 0.76 to 0.74, and
retains the same H@10 of 0.93. This suggests that FedV-KGQA transfers well to
longer reasoning paths on the movie-domain benchmark, likely helped by its
structured relations and the enriched graph. PathQuestion, in contrast,
shows a more noticeable decline. BERT+TransE drops from 0.65 to 0.57 in MRR,
and DistilBERT+DistMult drops from 0.66 to 0.51. This larger degradation may
reflect the smaller training set and the more heterogeneous family and
biographical relation chains in Freebase13, which make 3-hop reasoning more
difficult under the Silo-3 partition. RoBERTa+TransE exhibits the largest drops on PathQuestion across Hits metrics (up to 0.22 in H@5), suggesting that its weaker 2-hop performance also carries over to longer reasoning paths. Nevertheless, FedV-KGQA handles 3-hop reasoning without any architectural modification: extending it to three hops requires only one additional bounded expansion hop and reuses the Phase~0 rule set unchanged.
We evaluate $L \geq 2$ because single-hop questions resolve within
one relation and never span silos. We stop at three hops because the benchmarks
provide no deeper splits, not because the framework is limited to that depth.

\begin{figure}[t]
    \centering
    \includegraphics[
        width=\linewidth,
        height=0.25\textheight,
        keepaspectratio
    ]{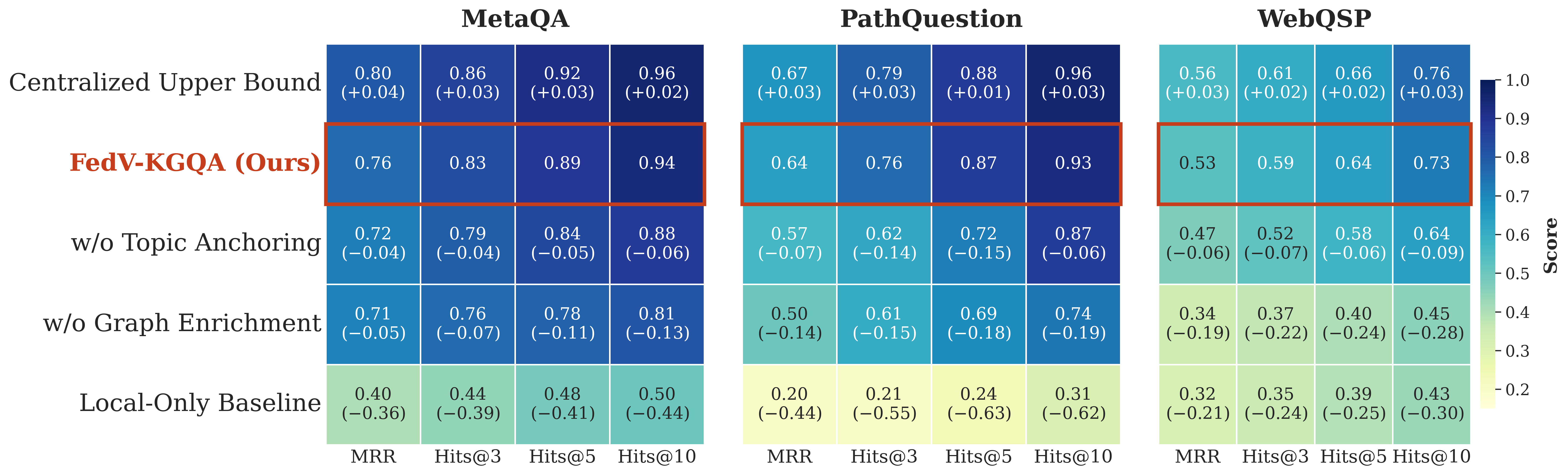}
    \caption{Ablation study across three datasets (deltas from FedV-KGQA in parentheses).}
    \label{fig:ablation}
\end{figure}

\begin{table}[t]
\centering
\caption{Comparison with adapted baselines. Best results are shown in \textbf{bold}.}
\label{tab:baselines}
\renewcommand{\arraystretch}{0.96}
\scriptsize
\resizebox{\textwidth}{!}{%
\begin{tabular}{l|cccc|cccc|cccc}
\hline
\multirow{2}{*}{\textbf{Method}} &
\multicolumn{4}{c|}{\textbf{MetaQA}} &
\multicolumn{4}{c|}{\textbf{PathQuestion}} &
\multicolumn{4}{c}{\textbf{WebQSP}} \\
& MRR & H@3 & H@5 & H@10
& MRR & H@3 & H@5 & H@10
& MRR & H@3 & H@5 & H@10 \\
\hline
Adapted EmbedKGQA~\cite{saxena2020improving}
& 0.26 & 0.27 & 0.30 & 0.35
& 0.32 & 0.36 & 0.44 & 0.51
& 0.30 & 0.34 & 0.37 & 0.42 \\
Adapted FL-KG-QA~\cite{gunti2025federated}
& 0.45 & 0.47 & 0.48 & 0.48
& 0.31 & 0.34 & 0.36 & 0.42
& 0.29 & 0.32 & 0.34 & 0.37 \\
Adapted FedE~\cite{chen2021fede}
& 0.70 & 0.77 & 0.83 & 0.88
& 0.58 & 0.68 & 0.78 & 0.87
& 0.46 & 0.51 & 0.57 & 0.62 \\
Adapted RelChain~\cite{jin2023improving}
& 0.71 & 0.78 & 0.83 & 0.88
& 0.57 & 0.65 & 0.74 & 0.88
& 0.42 & 0.48 & 0.53 & 0.60 \\
\hline
\textbf{FedV-KGQA (Ours)}
& \textbf{0.76} & \textbf{0.83} & \textbf{0.88} & \textbf{0.93}
& \textbf{0.65} & \textbf{0.81} & \textbf{0.89} & \textbf{0.96}
& \textbf{0.54} & \textbf{0.59} & \textbf{0.64} & \textbf{0.71} \\
\hline
\end{tabular}}
\end{table}

\subsection{ RQ3: Ablation Study}
Figure~\ref{fig:ablation} presents an ablation study using the
DistilBERT+TransE configuration with three silos across all three datasets.
We compare FedV-KGQA against four variants: a centralized upper bound that
trains on the merged KG without any federation, a version without
topic anchoring, a version without local graph enrichment, and a local-only baseline
where each silo answers questions independently using only its own embeddings. FedV-KGQA stays close to the centralized upper bound, trailing by 0.04 MRR on
MetaQA and 0.03 on PathQuestion and WebQSP. Removing topic anchoring drops MRR
consistently, most sharply on PathQuestion, where it falls from 0.64 to 0.57, and
on WebQSP, where H@10 drops by 0.09, confirming that grounding the question at the
topic entity is essential for accurate ranking. Disabling local graph enrichment
costs more, particularly on WebQSP where MRR falls by 0.19 and H@10 by 0.28,
because without inverse and chain axioms entity embeddings lose the bidirectional
training signal. The local-only baseline degrades most, with MRR falling to 0.40,
0.20, and 0.32, showing that no single silo holds enough relational knowledge to
answer multi-hop questions alone, and that federated fusion recovers performance
that would otherwise require centralized access to the full KG.

\subsection{RQ4: Baseline Comparison}
Since no existing method directly addresses multi-hop KGQA in the VFL setting, we adapt four representative methods to our VFL setup. For a fair comparison,
all methods use the same BERT+TransE configuration, KGE checkpoints, and 3-silo
partition. No original configuration exists to report, since each method assumes
centralized access, horizontal federation, or relation aggregation and does not
run unmodified in this setting. We exclude
LLM-based KGQA baselines because they require an additional retrieval and
prompting design over federated graph evidence, which is outside our KGE-based
VFL evaluation scope. None of the baselines include topic entity anchoring, which is
specific to FedV-KGQA. Table~\ref{tab:baselines} presents the results. EmbedKGQA~\cite{saxena2020improving} and
FL-KG-QA~\cite{gunti2025federated} show the largest gaps, with MRR below
0.46 on all datasets. EmbedKGQA uses average pooling to fuse silo
embeddings, which collapses the distinct geometric structure that each
relation partition contributes. FL-KG-QA restricts candidates to one-hop
neighbors and therefore cannot reach answer entities that require two-hop
reasoning. FedE~\cite{chen2021fede} and RelChain~\cite{jin2023improving}
are more competitive, reaching MRR values of 0.70 and 0.71 on MetaQA, respectively. However, FedE averages entity embeddings across silos after
each training epoch, which replaces the distinct representations learned
by each silo with a single shared embedding, losing the relation-specific
information that each partition contributes. RelChain performs comparably
to FedE on MetaQA and PathQuestion but drops to an MRR of 0.42 on WebQSP,
where its chain predictor struggles with the diverse open-domain relation
types. FedV-KGQA outperforms all baselines across all three datasets,
confirming that concatenation-based fusion and topic entity anchoring
provide a more effective foundation for multi-hop reasoning in the
VFL setting.

\begin{figure}[t]
    \centering
    \includegraphics[
        width=\linewidth,
        height=0.24\textheight,
        keepaspectratio
    ]{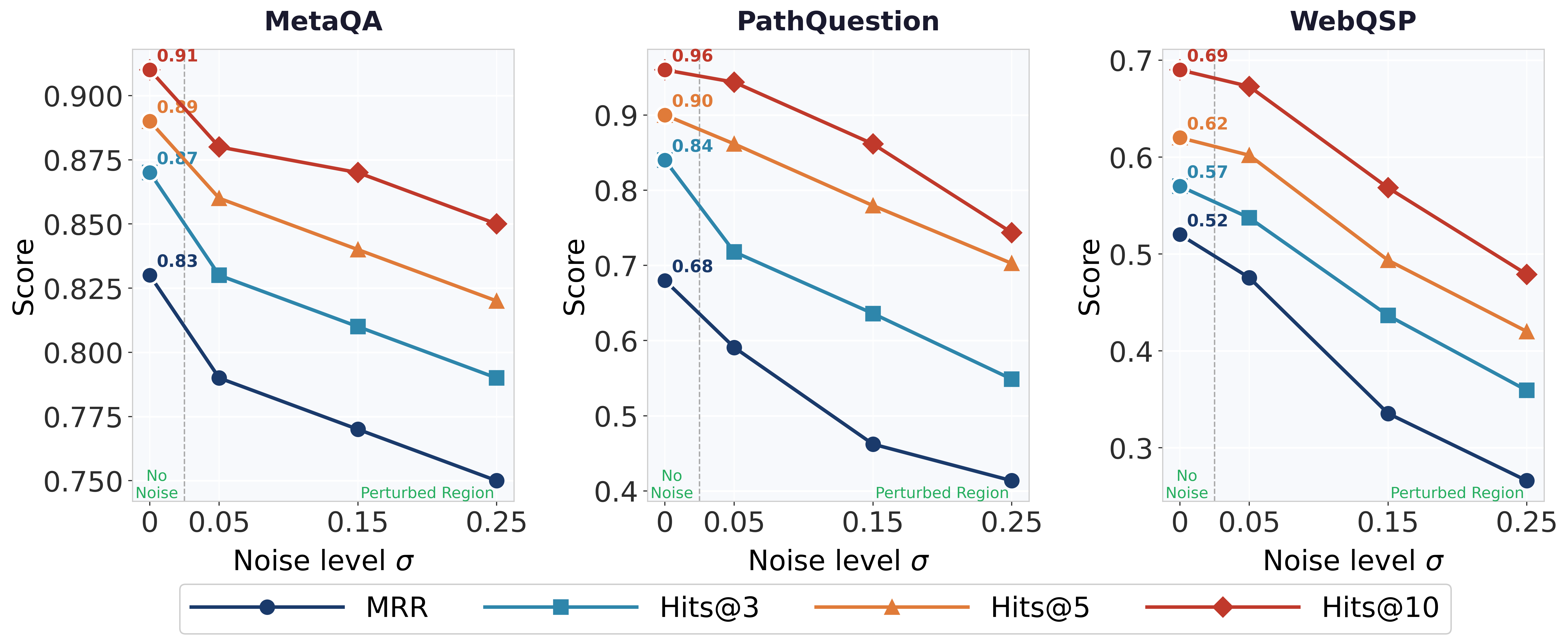}
    \caption{FedV-KGQA robustness to Gaussian noise (BERT+TransE, five silos). 
$\sigma = 0$ denotes the no-noise baseline.}
    \label{fig:dp-robustness}
\end{figure}

\begin{figure}[t]
    \centering
    \includegraphics[
        width=\linewidth,
        height=0.22\textheight,
        keepaspectratio
    ]{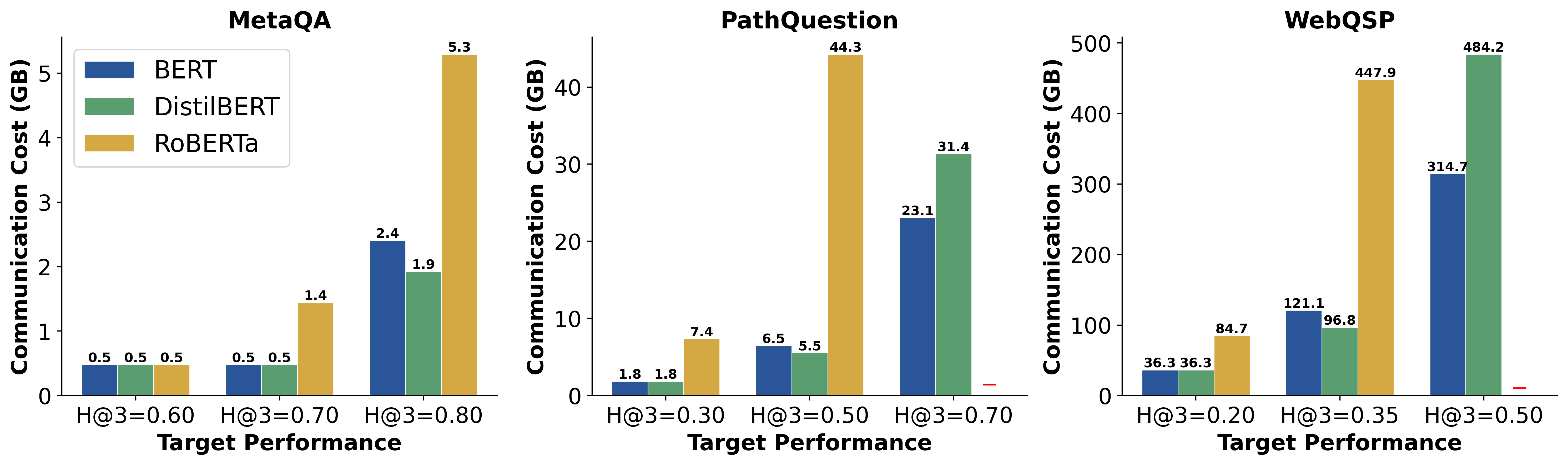}
    \caption{Communication cost in GB to reach H@3 targets across encoders (TransE, three silos). A dash (—) indicates the target was not reached.}
    \label{fig:comm-cost}
\end{figure}

\subsection{RQ5: Robustness to Embedding Perturbations}

Figure~\ref{fig:dp-robustness}  examines the robustness of FedV-KGQA to Gaussian noise 
$\mathcal{N}(0, \sigma)$ injected into entity embeddings, simulating  noisy communication channels in federated deployments. At low noise levels
($\sigma \leq 0.05$), performance remains largely stable across all three
datasets. On MetaQA, MRR decreases from 0.83 to 0.79 and H@10 from 0.91 to 0.88 at $\sigma{=}0.05$,
which is a modest decline. PathQuestion follows a
similar trend, with MRR dropping from 0.68 to 0.59 and H@10 from 0.96 to
0.94 at $\sigma{=}0.05$. WebQSP also shows a moderate decline, with MRR
dropping from 0.52 to 0.48 and H@10 from 0.69 to 0.67 at $\sigma{=}0.05$.
As $\sigma$ increases beyond 0.10, all metrics decline more steeply,
indicating that stronger noise levels come at a measurable cost to QA accuracy. Notably, even at $\sigma{=}0.15$, MetaQA retains an H@10 of 0.87
and PathQuestion retains an H@10 of 0.86, suggesting that FedV-KGQA retains
a degree of robustness under moderate embedding perturbations.

\subsection{RQ6: Communication Efficiency}
Figure~\ref{fig:comm-cost} shows the estimated communication cost to reach three H@3 targets for each dataset using TransE with three silos. Each silo uploads its entity embedding matrix and receives the corresponding gradient slice, with the cost per epoch given by $C_{\text{epoch}} = K \times |\mathcal{E}| \times d \times 2 \times 4$ bytes, where $K$ is the number of silos, $|\mathcal{E}|$ is the entity vocabulary size, and $d$ is the embedding dimension. The total cost is $C_{\text{total}} = T \times C_{\text{epoch}}$, where $T$ is the number of epochs. Cost therefore scales with the entity
vocabulary and is independent of how many triples each silo holds. DistilBERT is cheaper at moderate targets, requiring 96.8~GB against 121.1~GB
for BERT to reach H@3=0.35 on WebQSP, while BERT is cheaper at high targets,
needing 23.1~GB against 31.4~GB at H@3=0.70 on PathQuestion. RoBERTa is
consistently most expensive and misses all top targets.

\subsection{Summary of Findings}
TransE with BERT or DistilBERT is the most stable configuration, and
complex-valued models gain nothing from their larger dimension budget (RQ1). The framework extends to 3-hop reasoning without architectural
change, at a small cost on MetaQA and a larger one on PathQuestion (RQ2). Every
component contributes, with graph enrichment the largest single factor, the
local-only baseline confirming that no silo answers multi-hop questions alone,
and FedV-KGQA staying within 0.03 to 0.04 MRR of the centralized upper bound
(RQ3). FedV-KGQA leads all adapted baselines (RQ4). Accuracy degrades gracefully
under moderate embedding perturbation (RQ5), and communication cost is dominated
by entity count rather than local triple count (RQ6).

\section{Conclusion}
\label{sec:conclusion}

We introduced FedV-KGQA, a framework for multi-hop KGQA over
vertically partitioned knowledge graphs. Real-world knowledge is rarely
owned by a single organization, yet existing KGQA methods assume
centralized graph access. FedV-KGQA addresses this gap by combining local
KGE training, server-side embedding fusion, and topic entity anchoring to
enable cross-silo reasoning without sharing raw triples or relation
parameters. Experiments on MetaQA, WebQSP, and PathQuestion
show strong performance across 12 model configurations, with the framework
generalizing to 3-hop reasoning and remaining robust under embedding
perturbations. These results demonstrate that effective multi-hop question
answering is achievable even when the knowledge graph is split across
organizations. The key limitations are the assumption of a static graph and the absence of formal
DP guarantees for the embedding exchange protocol. We also do not measure how much of a silo's local structure could be recovered
from $\mathbf{H}_k$. Embedding inversion attacks are the natural test of the
structural boundary. Future work will focus on supporting incremental triple updates,
quantifying that leakage, and integrating DP mechanisms such
as noise calibration and secure aggregation.

\newpage

\subsection*{Supplemental Material Statement} 

The source code and interactive demo are available under the Apache License 2.0:

\begin{itemize}
    \item Source code: \url{https://github.com/brains-group/fedv-kgqa-source}
    \item Interactive Demo: \url{https://github.com/brains-group/fedv-kgqa-prototype}
\end{itemize}

\subsection*{Declaration of use of Generative AI}
Claude (Anthropic) was utilized to assist in the preparation of this work,
specifically for polishing and improving the writing of the manuscript and
for assisting with code development. All content, results, and conclusions
remain the sole responsibility of the authors. No AI tool was used to generate research findings, tables, or citations. 

%
%
%
\bibliographystyle{splncs04}
\bibliography{mybibliography}

\begin{thebibliography}{8}
\bibitem{ref_article1}
Author, F.: Article title. Journal \textbf{2}(5), 99--110 (2016)

\bibitem{ref_lncs1}
Author, F., Author, S.: Title of a proceedings paper. In: Editor,
F., Editor, S. (eds.) CONFERENCE 2016, LNCS, vol. 9999, pp. 1--13.
Springer, Heidelberg (2016). \doi{10.10007/1234567890}

\bibitem{ref_book1}
Author, F., Author, S., Author, T.: Book title. 2nd edn. Publisher,
Location (1999)

\bibitem{ref_proc1}
Author, A.-B.: Contribution title. In: 9th International Proceedings
on Proceedings, pp. 1--2. Publisher, Location (2010)

\bibitem{ref_url1}
LNCS Homepage, \url{http://www.springer.com/lncs}, last accessed 2023/10/25
\end{thebibliography}
%

\end{document}